\documentclass{llncs}
 
\usepackage[utf8]{inputenc}
\usepackage[T1]{fontenc}

\usepackage{color}
\usepackage{xcolor}
\usepackage{commath}

\usepackage{graphicx} 
\usepackage{amsmath} 
\usepackage{amssymb}  
\usepackage{verbatim}
\usepackage{algorithm}
\usepackage{algpseudocode}
\usepackage{tikz}
\usetikzlibrary{positioning}

\usepackage[inline]{enumitem} 

\usepackage{comment}
\usepackage{xcolor}
\usepackage{xspace}
\usepackage[normalem]{ulem}

\definecolor{OliveGreen}{RGB}{0,200,25}
\newcommand{\red}[1]{\textcolor{red}{#1}}

\newcommand{\darkgreen}[1]{\textcolor{OliveGreen}{#1}}

\newcommand{\eg}{e.\,g.\ }

\newcommand{\added}[1]{\blue{#1}}
\newcommand{\replaced}[2]{\red{\ifmmode\text{\sout{\ensuremath{#1}}}\else\sout{#1}\fi}\darkgreen{#2}}
\newcommand{\removed}[1]{\red{\ifmmode\text{\sout{\ensuremath{#1}}}\else\sout{#1}\fi}}

	\renewcommand{\added}[1]{#1}
	\renewcommand{\replaced}[2]{#2}
	\renewcommand{\removed}[1]{}
	
\newcommand{\removedfootnote}[1]{\footnote{\removed{#1}}}
\newcommand{\removedsubsection}[1]{\subsection{\texorpdfstring{\removed{#1}}{#1}}}

\title{\LARGE \bf
Uncertainty Estimation for Safe Human-Robot Collaboration using Conservation Measures
}

\author{Woo-Jeong Baek, Christoph Ledermann, and Torsten Kröger
\institute {Karlsruhe Institute of Technology (KIT) 
Institute for Anthropomatics and Robotics - Intelligent Process Automation and Robotics (IAR-IPR)}
\thanks{{\tt\small \{baek, christoph.ledermann, torsten\}@kit.edu}.}
\thanks{The authors would like to thank Tamim Asfour for his guidance and fruitful discussions throughout this work.}
}

\begin{document}

\graphicspath{ {./figures/} }

\maketitle
\thispagestyle{empty}
\pagestyle{empty}


\begin{abstract}
We present an online and data-driven {\it uncertainty} quantification method to enable the development of safe human-robot collaboration applications. 
Safety and risk assessment of systems are strongly correlated with the accuracy of measurements: Distinctive parameters are often not directly accessible via known models and must therefore be measured. 
However, measurements generally suffer from uncertainties due to the limited performance of sensors, even unknown environmental disturbances, or humans.
In this work, we quantify these \textit{measurement uncertainties} by making use of \textit{conservation measures} which are quantitative, system specific properties that are constant over time, space, or other state space dimensions. 
The key idea of our method lies in the immediate data evaluation of incoming data during run-time referring to conservation equations. 
In particular, we estimate violations of a-priori known, domain specific conservation properties and consider them as the consequence of measurement uncertainties. 
We validate our method on a use case in the context of human-robot collaboration, thereby highlighting the importance of our contribution for the successful development of safe robot systems under real-world conditions, \eg, in industrial environments. 
In addition, we show how obtained uncertainty values can be directly mapped on arbitrary safety limits (e.g, ISO 13849) which allows to monitor the compliance with safety standards during run-time. 
\end{abstract}

\section{INTRODUCTION}
Human-Robot Collaboration (HRC) systems have gained much attention in recent years: while enabling human workers to fully focus on more complex tasks, simple and repetitive ones are supposed to be performed by the robot. 
However, realizing a shared work space is bound to assuring a \textit{safe} environment throughout the run time. 
Existing literature in this domain as \cite{Askarpour2016}, \cite{Wadekar2018}, \cite{Inam2018} focus on hazard identification methods in simulation or based on pre-defined system models. 
The central assumption behind these approaches is that the measurement uncertainty in real-world scenarios is negligibly small such that findings obtained through models can be transferred to real-world applications.  
We, in contrast, argue that ignoring the existence of measurement uncertainties might lead to misinterpretations: dangerous situations could be classified as safe due to underestimated probabilities for undesired shifts.
Specifically in HRC systems, the robot parameters are usually accessible via control options while human parameters such as the position must be determined via measurements. 
Here, the actual human position might be closer to the robot as provided by the corresponding measurement device, such that the distance between human and robot could be underestimated. 
Depending on the robot velocity, the human might not be able to safely avoid the collision. 
Reversely, knowing the actual amount of measurement uncertainties would allow to prevent risks by adjusting parameters accordingly. 
For instance, an automatic reduction of the robot velocity could be initiated when exceeding a pre-defined threshold in the human localization uncertainty.\\ 
In this work, we present an online and data-driven method to quantify measurement uncertainties by leveraging domain-specific knowledge. 
In particular, we propose to make use of so-called \textit{conservation properties} of the system which are measures that are constant over state space parameters. 
In the context of human joint position detection, one conservation property is given by the non-changing distance between human joints. 
Formulating equations representing these conservation properties, we estimate violations thereof by means of incoming data. 
As we view the origin of such violations as the consequence of measurement uncertainties, we quantify latter ones by treating the data statistically. \\ 
The scientific novelty of our framework is given by its online and data-driven character:
By immediately analyzing incoming data with respect to formulated conservation equations and mapping the obtained uncertainty on arbitrary safety requirements (e.g., ISO 13849), we provide the possibility to monitor the compliance with safety standards online.   
In order to demonstrate the applicability our approach, we perform experiments for human-pose detection. 
The tolerated uncertainty is calculated in the context of standard ISO 13849, which used for safety verification purposes in industrial robotic applications. 
After presenting state-of-the-art methods for uncertainty determination in HRC, we introduce our methodology in Section \ref{method}. 
To validate our approach, we apply it on data collected with the safety scanner SICK3000 and compare our findings with the uncertainty stated in the corresponding sensor data sheet. 
Further, we discuss how the uncertainty quantification performs without taking conservation equations into account. 
In the last step, we demonstrate how the obtained uncertainty value can be mapped on safety requirements and thereby discuss the importance and limitations of our approach for safe HRC.

\begin{figure}[t]
     \centering
     \includegraphics[scale=0.2] {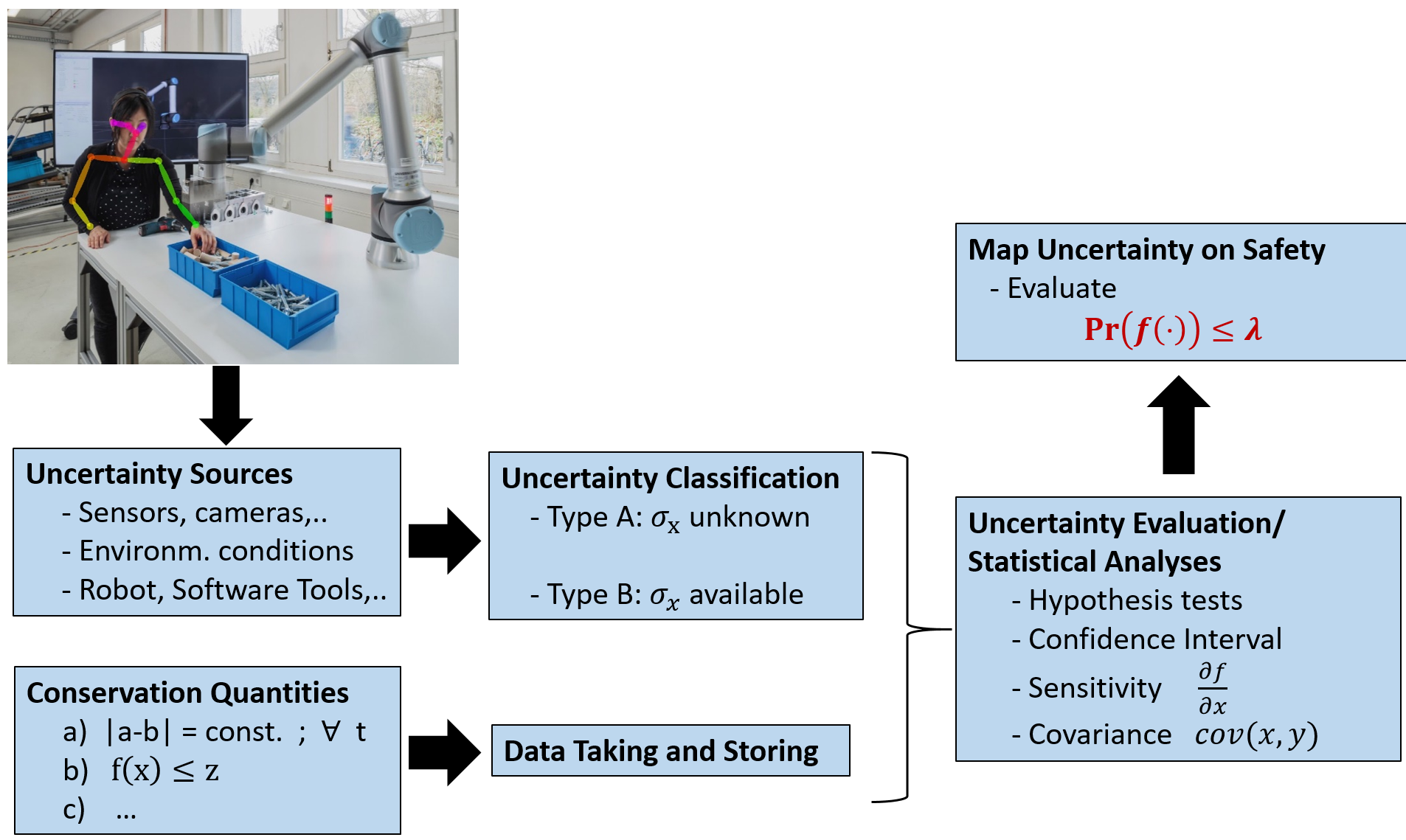}
     \caption{Combining the knowledge of uncertainty sources and conservation properties of a robot system with measurement data allows to effectively quantify uncertainties. 
     Treating obtained data with statistical tools, existing dependencies are explored. 
     Results are discussed in the context of safety compliance by referring to user-defined limits (e.g., ISO standards).}
\end{figure}

\section{Uncertainty and Safety in HRC} 
\label{SoTa} 
One of the main challenges in attaining safe HRC systems lies in steadily satisfying pre-defined limits. 
However, as highlighted in \cite{Hanna2020}, current safety engineering approaches are inadequate for flexible HRC. 
The authors state that existing safety assurance methods are well suited for traditional automation systems, but cannot be applied for intelligent or adaptive collaborative setups. 
In this work, we address this issue by presenting an online uncertainty quantification method. 
We claim that providing an uncertainty value throughout run-time is a crucial step towards flexible and safe HRC. 
For instance, the study of Bartneck et al. in ~\cite{Bartneck2020} highlights the necessity of an interpretable value for human beings to communicate the risk level. 
Contributions in the domain of safe HRC as \cite{Askarpour2016}, \cite{Inam2018}, \cite{Wadekar2018}, \cite{Gopinath2017} present hazard identification methods by modeling the system beforehand. 
While Askarpour et al. in \cite{Askarpour2016} present a formal verification method build upon a logic language to detect possible risks of the system based on its model, Inam et al. search for potential hazards in \cite{Inam2018} and generate a list of such by referring to a simulation setup. 
Similarly, Araiza et al. in \cite{Araiza2016} present a simulation-based verification method by developing a test generation approach based on different system variables. 
However, the authors of these contributions assume employed models to be accurate as they form the basis for suggested search strategies. 
In fact, uncertainties arising from the measurement process in real world environments are not taken into account. 
The survey in \cite{Lasota2014} discusses additional methods addressing the development of safe HRC systems, reaching from motion planning approaches to the consideration of psychological aspects. 
However, in contrast to our contribution, above works strongly focus on developing active search methods for possible risks.\\
In terms of human-involved systems, existing approaches as \cite{Haddadin2012}, \cite{Haddadin2008}, \cite{Nokata2002} and \cite{Ikuta2003} address the question how the knowledge on human injuries can be embedded in robot control frameworks to avoid critical harms on the human being. 
To do so, these works refer to parameters as the critical impact force and stress on different human body parts to evaluate whether HRC systems are safe or must be adapted with respect to control and design strategies to meet pre-defined safety requirements. 
One basic assumption of these contributions however is the accuracy of critical parameters which are decisive for decision-making. 
Therefore, uncertainties due to the measurement process the critical force and stress are not taken into account. 
In terms of robot applications, several contributions focus on the consideration of uncertainties in motion planning. 
In \cite{Thrun2006}, Thrun et al. introduce a variety of techniques to capture uncertainties originating from robot kinematics and to incorporate them in control frameworks.  
By presenting different representations of uncertainties, it is discussed how manipulation tasks can be accomplished in the face of uncertainties. 
Highlighting the need of online uncertainty tracking, several contributions as \cite{Nguyen2019}, \cite{Wirnshofer2018}, \cite{Liu2015}, \cite{Rahman2021} suggest to treat robot manipulation tasks in probabilistic manner.
Specifically, the authors present methods where robot tasks are modeled in the belief space. 
Here, each state is represented by a tuple of a state and the corresponding control input. 
By estimating subsequent states with a Bayesian framework, where the knowledge about the system state is combined with measurement updates, above works address the development of planning algorithms by taking robot uncertainties into account. 
Although suggested methods treat parameters in probabilistic manner, uncertainties of measurement updates originating from sensor devices are not explored in detail. 
To be specific, the authors note that the formulation of measurement updates considering possible uncertainties is challenging for unstructured environments:
Uncertainties due to sensors and measurement devices as well as their correlations with each other are either completely neglected or modeled as Gauss distributions. 
This assumption however is not applicable in all cases: In particular, the accumulated uncertainty resulting from the interplay of different sources might show a more complex behavior. 
In the context of safety, such incorrect assumptions could affect the decision making process in dangerous manner such that accurately quantifying uncertainties becomes highly crucial.\\
Generally, the contribution of Giancola et al. in \cite{Giancola2018} shows the highest similarity to our work: 
By presenting methods to determine the uncertainty behavior of three camera devices, the influence of environmental disturbances and system parameters is discussed in detail. 
Obtained results on correlations allow the user to adjust system parameters accordingly and to thus keep the uncertainty on a desired level.\\ 
Motivated by the fact that in existing literature, scarce attention is devoted to explicitly quantifying measurement uncertainties, we develop a framework which enables their online quantification during run-time. 
Furthermore, our method allows to map the uncertainty values directly on arbitrary safety-limits (e.g., ISO 13849). 
We adapt terminologies and methods introduced in \cite{GUM2009}. 
A detailed overview of definitions and analysis techniques will be given in Section \ref{method}. 
We apply our method on human position detection in a real-world environment to estimate the probability for the occurrence of dangerous failures per hour (PFDH), a distinctive measure stated in ISO 13849\cite{ISO13849}. 
We thereby \replaced{target}{address} following research questions:
\begin{enumerate*}[label=(\arabic*)]
    \item to which degree does our method provide accurate uncertainty results?, and  
    \item how can we quantify uncertainties of black box tools in the context of safety?
\end{enumerate*}
To the best of our knowledge, this work is the first contribution in this field that
\begin{enumerate*}[label=(\arabic*)]
    \item tackles measurement uncertainty quantification and propagation to evaluate robotic applications with respect to functional safety requirements, and 
    \item discusses how measurement uncertainties could contribute to the development of flexible and safe HRC systems.
\end{enumerate*}

\section{Methodology} \label{method}
\subsection{Problem Statement}
Generally, developing a safe system is related to satisfying an equation of the form
\begin{equation}
    Pr ( f(\cdot))\leq \lambda,
    \label{eq:1}
\end{equation}
where $Pr(\cdot)$ denotes the probability, $f(\cdot)$ a user-defined constraint and $\lambda$ an arbitrary limit (e.g., ISO 13849 \cite{ISO13849}).
At the same time, safety is achieved by minimizing the risks of the system. 
According to ISO 12100 \cite{ISO12100}, the term \textit{risk} can be written as
\begin{equation}
    risk_{system}=\sum_i severity(i)\cdot Pr(i).
\end{equation}
Here, \textit{i} stands for an incident and \textit{Pr(i)} for the probability of its occurrence.
In our work, we assume the severity, which can be for example biomechanical limits of the human's bones, as a constant and known quantity.
Thus, the risk minimization can be achieved by minimizing the probability:
\begin{equation}
    \min ({risk_{system}}) \equiv \min ({Pr(i)}).
\end{equation}
However, the probability for the occurrence of a dangerous outcome originates from the incomplete knowledge of the system behavior, specifically from the range of possible deviations from the expected outcome.
As indicated in Section \ref{SoTa}, the \textit{measurement uncertainty} $u$, which we will refer to with the term \textit{uncertainty} for the remainder of this work, covers these kind of deviations, i.e.:
\begin{equation}
    Pr(i) \propto u(i).
    \label{eq:prob_sigma}
\end{equation}
Reversely, the complete knowledge on the system with its possible deviations $u(\cdot)$ from the expected behavior would allow to predict and thus incorporate measures to avoid all dangerous situations beforehand.\\
Therefore, the development of a safe system requires a thorough analysis of its uncertainties.  
One main challenge in the uncertainty analysis is the identification of its dependencies on various parameters as time, space and environmental conditions: 
\begin{equation}
    u_{system}=f(u_1(s,t,...), u_2(s,t,...),..., u_n(s,t,...)),
    \label{eq:sigma_dep}
\end{equation}
where the system uncertainty $u_{system}$ is a function of the system components' uncertainties $u_j\; ; j \in [1;n]$.\\
In particular, the system uncertainty is always related to a specific property which we denote as \textit{attribute a}. 
For instance, in safety-critical applications, such an attribute might be the distance between human and robot, representing a critical quantity. 
The obtained result on the uncertainty would reflect the possible deviation on the distance. 
Generally, the goal is to quantify the total uncertainty which is induced by the system on an arbitrary, user-selected attribute. 
Hence, \replaced{at the end of}{in} our analysis, we provide a result of the form 
\begin{equation}
    a \pm u_{system,a}
\end{equation}

\subsection{Classification of Uncertainties}
To approach the issue of uncertainty estimation, we first need a proper classification thereof. 
As stated in Section \ref{SoTa}, we apply the terminology introduced in \cite{GUM2009}. 
According to this guide, the so-called \textit{input quantities} $x_{a,i}$ on attribute $a$ are classified regarding two evaluation methods: 
\begin{itemize}
    \item Frequency based uncertainty evaluation based on series of observations of the input quantities $x_{a,i}$ and subsequent statistical analyses (Type A)
    \item according to a-priori knowledge on the uncertainty behavior of input quantity $x_{a,i}$, that is provided by a theoretical model or the manufacturer (Type B).
\end{itemize}
In the following, we denote the uncertainty of input quantity $x_{a,i}$ with $u_{x_{a,i}}$.

\subsection{Uncertainty Quantification}
To properly determine the uncertainty on the attribute, its functional relationship with input quantities $x_{a,i}$ 
\begin{equation}
    a = g(x_{a,1},x_{a,2},...x_{a,m}); \; \; m \leq n
\label{eq:attribute}
\end{equation}
must be known beforehand. 
Here, the input quantities represent the measurands of components $c_j$. 
Specifically, the variable $x_{a,i}$ stands for the input quantity of component $c_i$, where $i \in [1;m]$ with $m \leq n$.
Above relationship which shows the dependency of attribute $\textit{a}$ on the system components $c_j$ with $j \in [1;n]$ is required to study the uncertainty propagation: 
The influence of an input quantity on the attribute determines in which sense its measurement is affected by the considered system component. 
Assuming the independence of measurement tools, the propagation results in a so-called \textit{combined standard uncertainty} $u_C(a)$ of 
\begin{equation}
    u_C(a)=\sqrt{\sum_j\abs{\frac{\partial a}{\partial x_{a,j}}} u(c_j)}
    \label{eq:uncert_prop}
\end{equation}
for attribute $a$. 
However, the explicit form of Eq.~\eqref{eq:attribute}  might not be available. 
In particular, the input quantities directly depend on the measurement uncertainty of employed devices such that the amount of $\frac{\partial a}{\partial x_{a,j}}$ is not directly quantifiable on the basis of theoretical knowledge. 
Hence, we aim to determine it via measurements. 
To do so, we leverage the knowledge of certain conservation properties which allows us to quantify the uncertainty originating from a measurement device. 
We apply common statistical tools including bootstrapping, hypothesis testing and correlation analyses. 
Since these analyses highly depend on the specific form of Eq.~\eqref{eq:attribute}, we discuss the uncertainty quantification on a specified use case in HRC. 

  \begin{figure*}[htbp!]
      \centering
      \includegraphics[width=0.9\textwidth]{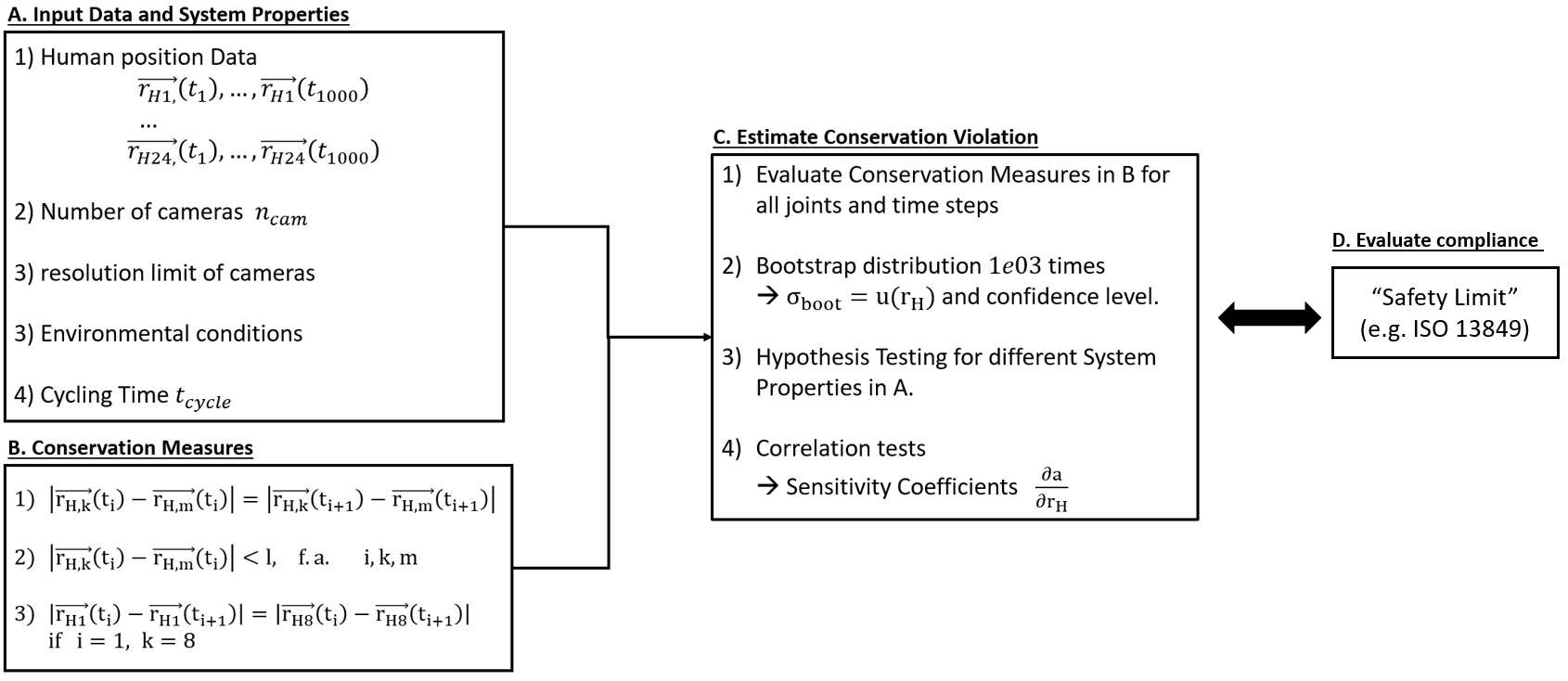}
      \caption{We leverage the knowledge of non-changing parameters in the state space for uncertainty quantification. Regarding violations on the conservation of these non-changing properties by means of incoming data, we determine the uncertainty online. In the last step, we map  the obtained uncertainty result on an arbitrary safety limit.}
      \label{fig_method}
   \end{figure*}

\subsection{Use Case: Human Robot Collaboration} \label{method_HRC}
In Section \ref{SoTa}, we outlined the relationship between uncertainties and safety assurance.
Functional safety of HRC setups depends on relative measures such as \textit{distance, velocity} and \textit{force} between human and robot. 
To allow for a reliable assessment whether a situation is critical or safe, these parameters must be measured with a high degree of accuracy. 
On the other hand, the accuracy is limited by measurement uncertainties which are often not understood in detail. 
Therefore, estimating their influence on above distinctive measures is an essential step towards safe HRC. 
Specifying Eq.~\eqref{eq:1} for the distance $d$ between human and robot yields
\begin{equation}
    Pr(|\Vec{r}_H-\Vec{r}_R|\leq d) \leq \lambda_{HRC} \; \; ; \;\;\;\;\;\; \forall t
    \label{eq:HRC}
\end{equation}
where $\Vec{r}_H, \Vec{r}_R$ stand for the positions in 3D-Cartesian space, respectively.
The safety assessment generally is achieved by evaluating Eq.~\eqref{eq:HRC}. 
It is obvious that this evaluation necessitates the knowledge of the robot and human position. 
While $\Vec{r}_R$ is easily accessible via control options, $\Vec{r}_H$ is challenging to capture. 
Since an accurate model representing the human behavior does not exist, its position is determined through measurements resulting in
\begin{equation}
    \tilde{\vec{r}}_H = \vec{r}_H \pm \vec{u}_H.
    \label{eq_relation}
\end{equation}
As a consequence, Eq.~\eqref{eq:HRC} changes to 
\begin{equation}
    Pr(|(\vec{r}_H \pm \vec{u}_H)-\Vec{r}_R|\leq d) \leq \lambda_{HRC} \; \; ; \;\;\;\;\;\; \forall t.
    \label{eq_HRCsigma}
\end{equation}
Thus, to evaluate Eq.~\eqref{eq_HRCsigma}, the knowledge of $\vec{u}_H$ is required.
In particular, its minimization is desired to allow for a high confidence level for the compliance with $\lambda_{HRC}$. 
In contrast to $\vec{u}_R$ which we can derive on the basis of manufacturer specifications, $\vec{u}_H$ reflects the detection uncertainty of the human movement which depends on employed tools. 
We target the determination of the combined standard uncertainty on the human position which can be formulated by
\begin{equation}
    u_C(\vec{r_H})=\sqrt{\abs{\frac{\partial r_H}{\partial det}} u(det) + \abs{\frac{\partial r_H}{\partial \rho}} u(\rho) + ...}.
    \label{eq:uncert_rH}
\end{equation}
Here, ${det}$ represents the detection performance of the human position and $\rho$ arbitrary environmental parameter such as lightning, temperature etc. 
We assume that applied devices are independent from each other, that is, the measurements of environmental parameters do not stand in any relation to the human position detection. 
As will be specified in Section~\ref{experiments}, we use the SICK safety scanner S3000 \cite{SICK3000} in addition to OpenPose \cite{Cao2019} for human localization. 
Consequently, the first term in above equation is split as follows: 
\begin{equation}
    \abs{\frac{\partial r_H}{\partial det}} u(det)= \abs{\frac{\partial r_H}{\partial det_{OP}}}u(det_{OP}) \cdot \abs{\frac{\partial r_H}{\partial det_{LS}}}u(det_{LS}).
\end{equation}
The safety scanner uncertainty $u(det_{LS})$ and such due to the behavior of environmental parameters $u(\rho)$ are assumed to be known and thus considered as Type B uncertainties.
In contrast, $u(det_{OP})$ arising from OpenPose is not known due to a missing understanding of neural network uncertainties to date. 
To this end, we categorize it as Type A uncertainty and conduct analyses as explained in the following.\\

The key idea of this work lies in the \textit{conservation based uncertainty estimation}, where conservation properties reflect characteristics of the system which can not change over state space parameters. 
Formulating conservation equations and evaluating them by means of incoming data during run-time allows to study the amount of violations. 
To be specific, conservation equations represent conditions on user-defined parameters which are known to stay constant.  
Given the knowledge that violations thereof can not occur, we view their occurrence as a direct consequence of the measurement uncertainty of the applied tool. 
Since the conservation equations are evaluated online, we obtain a distribution which reflects the amount of violations. 
Applying statistical techniques, we determine the measurement uncertainty and its confidence level.\\
In the use case of HRC, we focus on the determination of $\frac{\partial r_H}{\partial det,OP}$. 
One basic assumption of our work is the general functionality of the applied tool, that is, we do not focus on out-of-distribution events. 
Instead, we aim to quantify deviations within the scope of functionality and determine a confidence metric referring to measurement data. 
Here, we leverage the conservation of the constant distance between two human joints throughout time, that is
\begin{equation}
    \abs{\vec{r}_{H,j}(t) - \vec{r}_{H,k}(t)} = const. \; \; ; \;\;\;\; \forall t; \; \; j\neq k.
    \label{eq:constraint}
\end{equation}
In our analysis, we assume that all joints of the human body are detected with the same performance. 
For each of our datasets, we calculate the distance between two joint pairs of the human body and obtain distributions.  
After collecting data of ten time steps, we conduct following procedure for the uncertainty determination and the mapping of such on the user-chosen safety limit: \begin{enumerate}
    \item \textbf{Bootstrapping:} We perform random sampling with replacement from our dataset ten thousand times. 
    Each data point is assigned equal probability such that one data point can be selected several times. 
    For each distribution, the mean value is calculated. 
    According to the Central Limit Theorem, these mean values follow a Gaussian distribution which enables the uncertainty estimation on a confidence interval. 
    \item \textbf{Hypothesis Testing:} Hypothesis tests are performed on the bootstrapped distributions. 
    We state which dependencies are indicated, that is, whether the uncertainty behavior shows tendencies regarding parameters of interest. 
    \item \textbf{Covariance:} In case a dependency is indicated, we analyse the corresponding covariance. 
    Leveraging the knowledge of this measure allows the user to adapt the system accordingly by reducing the uncertainty. 
    \item \textbf{Mapping on customizable safety limit:} As stated above, we view uncertainties as the main cause for the probability of occurrence for dangerous situations. 
    Thus, we directly map the uncertainty on any safety limit:
    For instance, standard ISO 13849 states a maximum rate of $10^{-6}$ for the occurrence of dangerous situations per hour. 
\end{enumerate}

Generally, our method is applicable on arbitrary systems where conservation equations can be formulated and evaluated online according to Algorithm \ref{alg:cap}. 
However, for most well-understood devices, the uncertainty information is accessible via manufacturer specifications. 
Therefore, our method becomes particularly interesting for cases, where the uncertainties are not known. 
We view OpenPose as one measurement tool for the detection of human joint positions with a complex and non-obvious uncertainty behavior. 
This motivates us to apply our method on OpenPose to explore its uncertainty.

\begin{algorithm}
\renewcommand{\algorithmicrequire}{\textbf{Input:}}
\renewcommand{\algorithmicensure}{\textbf{Output:}}
\caption{Conservation based Uncertainty Estimation}\label{alg:cap}
\begin{algorithmic}
\Require measurement data $x_{a,i}$; conservation Eq. $f_C(\cdot)$; \\
relationship betw. attribute and data $a(x_{a,i})$; confidence level $\sigma$; parameter for correl. analysis $\xi$; safety limit $\lambda$; time steps $t$.
\Ensure total combined uncertainty $u_C(a)$; \\safety limit check (bool)
\For{$i \leq t$}
\State $dev[i] \gets f_C(x_{a,i},...,x_{n,i})$
\EndFor

\For {$z \leq 10.000$} 
\State $b[z] \gets $ bootstrap dev[]
\EndFor \\
From b[] compute $u_C(a)$ for user-defined $\sigma$\\
Test $H_0$ for given p-value and parameter $\xi$

\If{$H_0$ rejected}
    \State $cov(u_C, \xi)$
    \State print $cov(u_C,\xi)$
\EndIf \\
$r \gets u_C \cdot l_{bio}$
\If{$r\leq \lambda$}
\State return 1
\Else
\State return 0
\EndIf
\end{algorithmic}
\end{algorithm}


\section{Experiments} \label{experiments}
\begin{figure}[htbp] 
     \centering
     \includegraphics[scale=0.05] {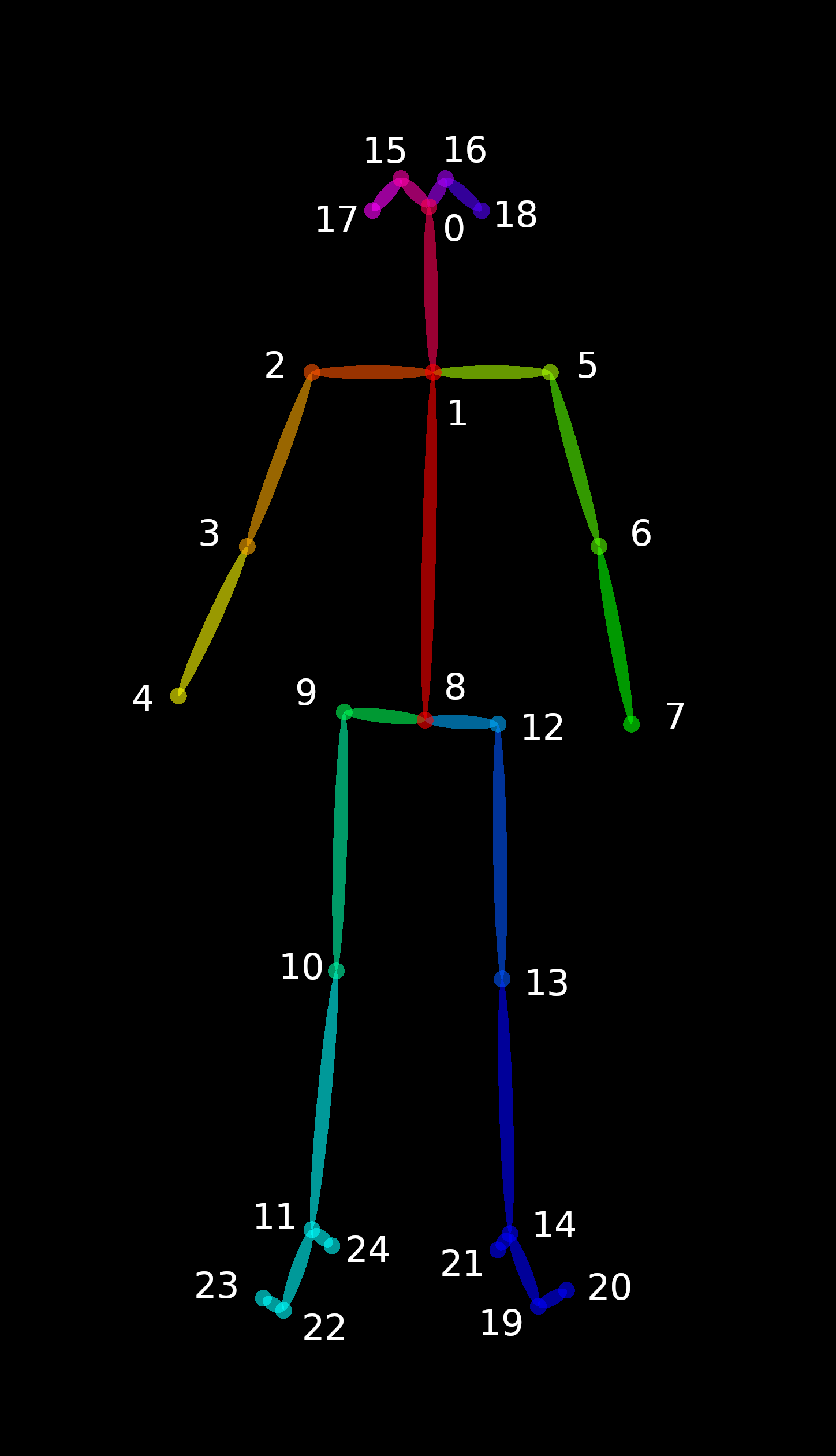}
     \caption{We apply the model of Body25 to detect the key points with OpenPose.}
     \label{fig_body25}
\end{figure}

To validate our method, we refer to the SICK safety scanner. 
Since the data sheet of the safety scanner provides an uncertainty value, it offers the possibility to assess the accuracy of our approach. 
In the second step, we address the uncertainty analysis of OpenPose. 
As will be detailed in the following, we make use of datasets that provide ground truth information. 
Treating OpenPose as a measurement device for human joint localization, we formulate conservation equations upon domain specific knowledge and quantify the uncertainty by applying our method.  
We consider the human body model shown in Fig.~\ref{fig_body25} and perform our analyses by means of following datasets: 

\subsection{Datasets}
\begin{enumerate}
    \item \replaced{Data Collection 1}{Dataset 1:} Action Recognition NTU RGB+D is a large scale data set which contains markerless human movement data. 
    The published ground truth data for the joint positions was obtained with Microsoft Kinect V2. 
    Each skeleton is represented by 25 joints. 
    As the uncertainty of Microsoft Kinect V2 has been studied in detail in \cite{Giancola2018}, we consider findings in our analysis respectively. Thus, we treat the ground truth values as distributions to represent according deviations. 
    \item  \replaced{Data Collection 2}{Dataset 2:}
    We have gathered RGB+D data of six moving humans in a HRC setup. 
    To obtain the ground truth for the conservation property in Eq.~\eqref{eq:constraint}, we measured the distance between joints manually. 
    Since this method is quite inaccurate, we consider an uncertainty of $5\,\%$ arising from the manual measurement process. 
    In addition, we refer to the positioning uncertainty of UR10e \added{robot} provided by Universal Robots (UR) to calculate the uncertainty on the $d$.
\end{enumerate}

\subsection{Conservation Properties and Type B Uncertainties}
We define the distance between the human joints as conservation properties as formulated in Eq.~\eqref{eq:constraint} and apply OpenPose to above datasets. 
In both cases, we estimate deviations with respect to the conservation equations online to explore the uncertainty of OpenPose. 
As a next step, we combine our results with the a-priori known uncertainty behavior of applied devices (Type~B uncertainties). 
In case of dataset 1, we refer to the uncertainty of Microsoft Kinect V2. 
As presented in \cite{Giancola2018}, the uncertainty behavior can be modeled by a linear function
\begin{equation} \label{eq:giancola_Kinect}
    u_{Kinect}(r_K)= 8\cdot10^{-4} \cdot r_K - 1\cdot 10^{-4},
\end{equation}
where $u_{Kinect}(r_K)$ stands for the Microsoft Kinect V2 uncertainty and $r_K$ for the distance to the camera. 
For dataset 2, in contrast, we refer to UR10e specifications given by
\begin{equation}
    u_{UR10e}=0.1\cdot10^{-3}\,\mathrm{m}.
    \label{eq_UR}
\end{equation}
In particular, we consider this value to determine the distance uncertainty between human and robot for each time step. 
We use a safety laser scanner as an additional device to track the human feet. 
Since its uncertainty is provided by the manufacturer SICK, it is treated as Type~B uncertainty. 
Applying our method on static data collected by the safety scanner allows us to assess its accuracy as will be discussed below.

\subsection{Validation, Evaluation and Results}
Prior to applying our method on the data, we conduct a validation thereof. 
Since the safety scanner uncertainty is provided by SICK, it enables us to compare the results obtained through our approach. 
We record static situations, that is, we perform our data collection in an environment with static obstacles and non-changing environmental conditions. 
The measurement data consists of data points collected at a scanning range of $4.0\,m$ covering an angular range of 275°, where each scanning interval amounts to 0.3850°. 
We follow the steps presented in Section \ref{method_HRC} for uncertainty calculation. 
  \begin{figure}[htbp]
      \centering
      \includegraphics[width=0.8\textwidth]{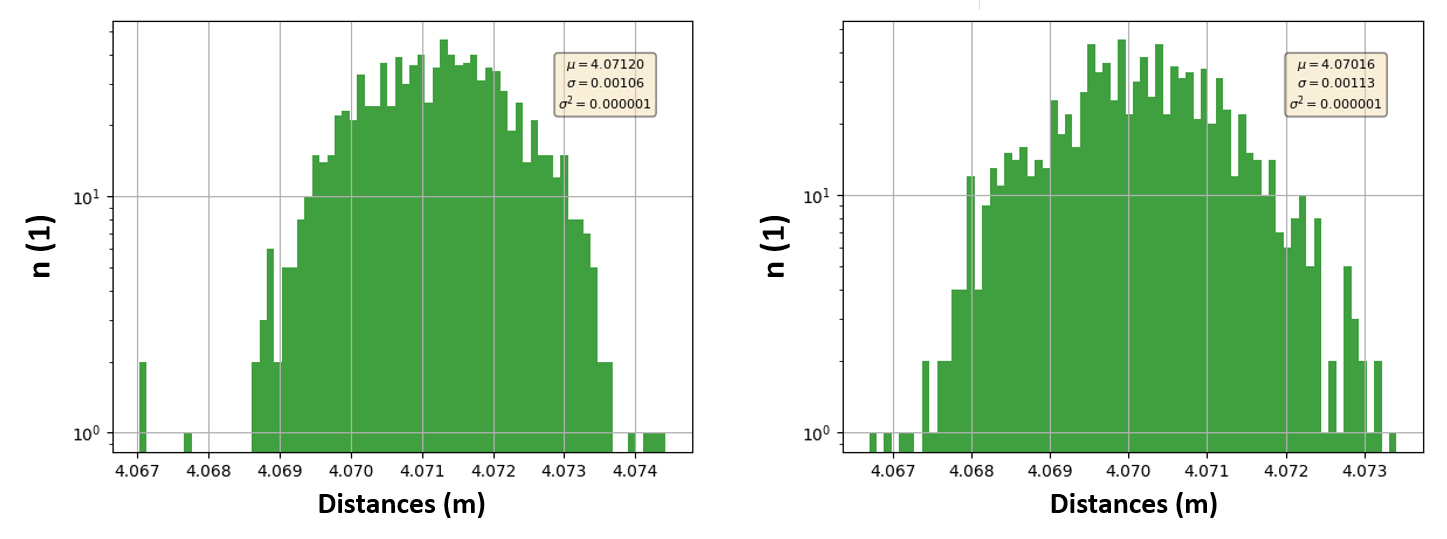}
      \caption{We estimate the uncertainty of SICK safety scanner 3000 to validate our approach. We assume a Gaussian behavior of the uncertainty and perform our calculation referring to 100 measurements of static situations.}
      \label{OL_static}
   \end{figure}
The bootstrapping procedure is carried out 10.000 times for each angular setting. 
Example histograms depicting the bootstrapped distributions are shown in Fig.~\ref{OL_static}.
According to our results, the systematic uncertainty of the SICK safety scanner S3000 amounts to $ u_{LS,exp}=0.0038\,m$ at a scanning range of $4.0\,m$ which equals the relative uncertainty of 
\begin{equation}
    u_{LS,exp}=0.095\,\%
\end{equation}
on a $95\%$ confidence level. 
The data sheet provided by SICK states an uncertainty of $u_{LS,theo} = 0.0050\,m $ at a scanning range of $5.5\,m$ which refers to a relative uncertainty of 
\begin{equation}
    u_{LS,theo} = 0.091\,\%.
\end{equation}
Thus, our method yields a result which differs approximately $4.0\%$ from the manufacturer specifications. 
In order to examine to which extent our method improves the uncertainty quantification, we compare this result with the baseline study: 
Hence, we determine the uncertainty by analysing the spread of obtained data from the laser scanner. 
As we recorded static situations, we do not expect any fluctuations in the measured values. 
However, without incorporating any conservation equations in our analysis, we obtain an uncertainty of $u_{LS,baseline}=0.0054\,m$ for the same data set. 
Hence, the baseline method without conservation properties yields a relative uncertainty of 
\begin{equation}
    u_{LS,baseline} = 0.14\,\%,
\end{equation}
which equals a difference of $30\,\%$ from the SICK data sheet.\\

To analyse the uncertainty of OpenPose, we evaluate conservation equations Eq.~\eqref{eq:constraint}, more specifically deviations thereof, online.  
In analogy to the safety scanner analysis, the resulting distribution allows to determine the uncertainty on a certain confidence level.
However, we are additionally interested in the relationship between the uncertainty and the human velocity.  
We therefore target the exploration of this correlation by performing statistical tests with respect to following null hypothesis: 
\begin{quote}
\textit{H0: The uncertainty does not increase with higher velocities in the human movement.}
\end{quote}
Our analysis focuses on certain human joints for which we assume no significant variations during the movements. 
To investigate dataset 1, we refer to uncertainty of Microsoft Kinect V2 stated in Eq.~\eqref{eq:giancola_Kinect}. 
Consequently, we obtain distributions for the ground truth positions. 
We apply OpenPose making use of the body-25 key point model and compare results with the ground truth data.
Regarding dataset 2, we additionally consider the uncertainty of Intel RealSense D435 as well as the detection performance of SICK S3000 (Type B uncertainties). 
Finally, we determine the total uncertainty on the distance between human and robot: 

\begin{equation}
    u_C(d_{HR}) = \sqrt{\abs{\frac{\partial d_{HR}}{\partial r_H}} \cdot u_C(r_H)+\abs{\frac{\partial d_{HR}}{\partial r_R}}\cdot u(r_R)},
\end{equation}
where $d_{HR}$ is given by

\begin{equation}
    d_{HR} = \sqrt{\sum_{p=1}^3 ({r_{H,p}-r_{R,p})^2}}
\end{equation}
since we conduct analyses for 3D human pose estimation.
We obtain following expression for the human-robot distance uncertainty :
\begin{equation}
    u_C(d_{HR})=  \frac{\sum_{p=1}^3(r_{H,p}-r_{R,p})}{d_{HR}}\cdot [u_c(r_H)+u(r_R)],
\end{equation}
where we use the manufacturer specifications for $u(r_R)$ defined in Eq.~\eqref{eq_UR}. 
Furthermore, we evaluate Eq.~\eqref{eq:uncert_rH} online during run-time. 
To explore the uncertainty behavior of OpenPose in dataset 2, we subtract the relative uncertainty of $\leq 2\%$ stated in the data sheet of Intel RealSense D435 from the total uncertainty. 
It is noted that the uncertainty arising from the calibration process of the cameras and the triangulation into 3D space is negligible. 

\begin{figure*}[htbp]
\centering
\includegraphics[width=0.8\textwidth]{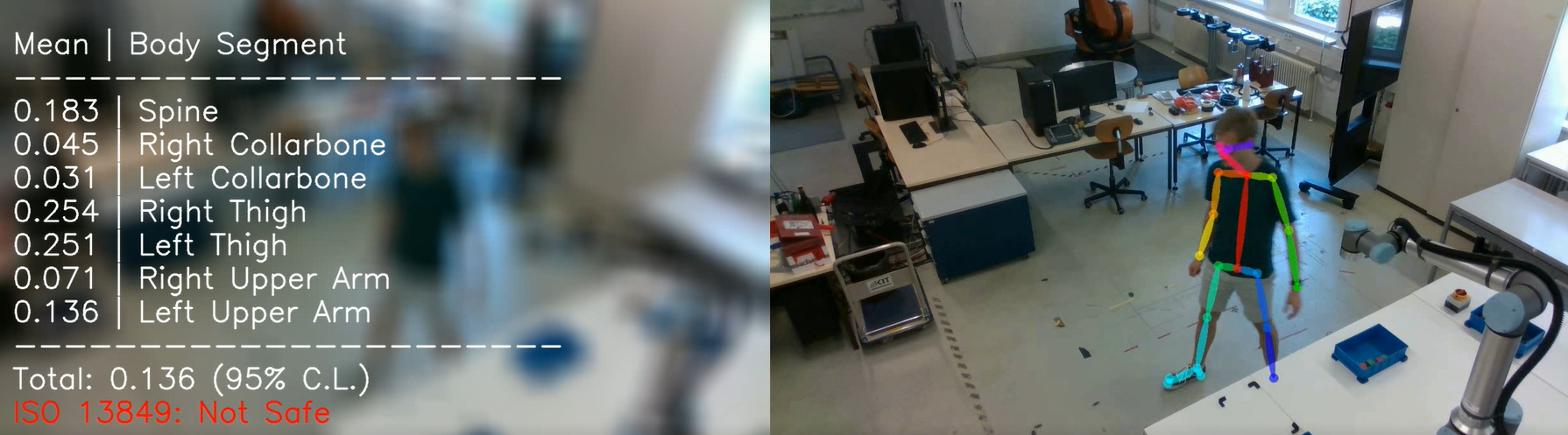}
\caption{Our method allows to estimate the uncertainties on user-selected joint pairs online. During run-time, our method provides uncertainty values and an evaluation on an arbitrary safety limit (e.g., ISO 13849).}
\label{video_static}
\end{figure*}

\subsection{Results}
\begin{figure}[htbp]
\centering
\includegraphics[scale=0.25]{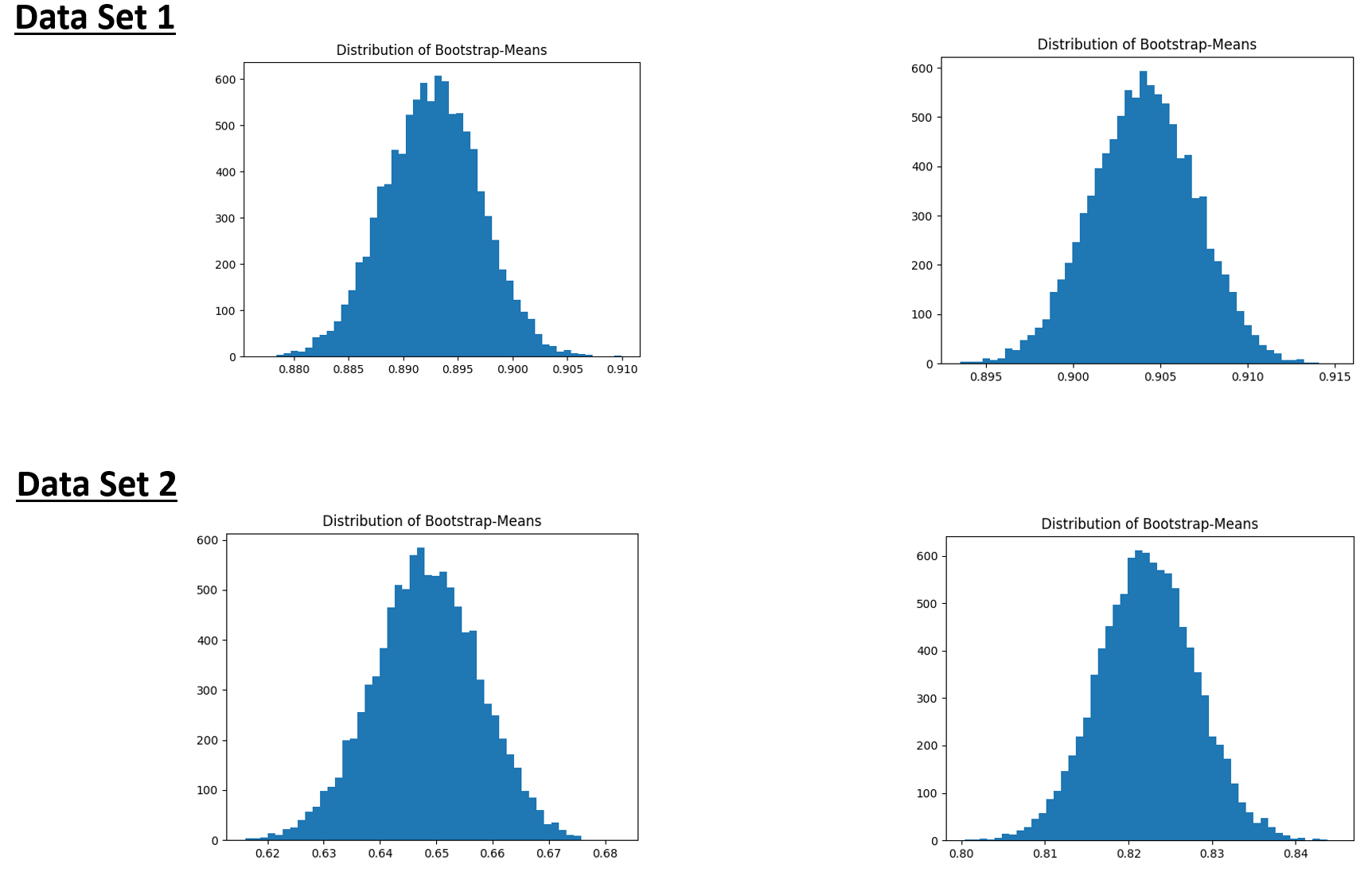}
\caption{Bootstrapping the deviations from the ground truth value of the OpenPose output 10.000 times yields Gaussian distributions. We show two example plots for each dataset.}
\label{fig:bootstrapped_distr}
\end{figure}

In our analysis, we focus on the joint pairs of the upper body part. 
As can be seen in Fig.~\ref{video_static}, the lower body part is often not detected, especially in the direct surrounding of the robot. 
We refer to joint pairs where minimal variations are expected. 
While the segment between the elbow and the hand is assumed to underlie more movements and thus more fluctuations, the part between the neck and hip is expected to show a more constant behavior. 
Two example histograms illustrating the bootstrapped distributions of the uncertainty are shown in Fig.~\ref{fig:bootstrapped_distr}.

Our method yields following results for the 3D spatial detection uncertainty of OpenPose:
\begin{equation}
    u_{D1}(det_{OP})=0.008\,\%
\end{equation}
\begin{equation}
    u_{D2}(det_{OP})=0.015\,\%. 
\end{equation}
Accordingly, the uncertainty estimated in \replaced{data collection 2}{dataset 2} is double as high as determined on dataset 1. 
It is very likely that this discrepancy is due to the lower statistical significance of dataset 2. 
While dataset 1 consists of a large data amount, we refer to data of six different human beings in dataset 2. 
Furthermore, the uncertainty of $u_{FR}=5\,\%$ which we assumed to originate from the manual measurement of the ground truth is one possible origin for above discrepancy. 
Our final result which we obtain through averaging both values is given by 
\begin{equation}
    u_{tot}(det_{OP})=0.009\,\%
\end{equation}
on a $95\%$ confidence level. 
Regarding the hypothesis test, above defined null hypothesis is rejected for both data collections with a p-value of $p=0.9816$. 
Conducting correlation analyses with the human velocity leads to an averaged correlation of $0.32$. 

\subsection{Mapping on Safety Limit}
Our approach allows to map the uncertainty on any user-defined limit. 
Since we are interested in robotic applications, we refer to the safety limit in ISO 13849 stating a maximum probability for the occurrence of dangerous failures per hour $\mathrm{PFD_H}$ of
\begin{equation}
    \mathrm{PFD_{H,max}} = 10^{-6}/h. 
\end{equation}
As we consider uncertainties as the main origin for the evolution of this probability, we directly map our results on this quantity. 
According to ISO 13849, one dangerous failure in $10^6$ hours is tolerated.
Above calculated uncertainty of $0.009\,\%$ applies to one dangerous failure per $10^4$ hours, assuming that slight uncertainties in the human joint position detection could yield critical situations. 
Fig.~\ref{video_static} shows a static depiction of our methodology. 

\subsection{Limitations}
Our experiments and validation demonstrated that our conservation based uncertainty quantification method yields reasonably accurate results. 
Nevertheless, our framework requires the formalization of conservation equations by the user which could cause challenges in applying it to more complex systems. 
For instance, identifying conservation measures which offer the possibility to be monitored during run-time might not always be straight-forward. 
In contrast to above presented scenarios, where the length between the human body joints could be evaluated directly by incoming data, conservation properties could be separated from the monitored data. 
In addition, more complex conservation equations might cause computational burden on our framework. 
As a consequence, the online uncertainty quantification could be affected. 
However, these possible shortcomings have not been studied within the frame of this contribution. 
We therefore suggest to perform more detailed studies thereof in future. 

\subsection{Discussion}
In order to validate the accuracy of our method, we applied it on static data recorded with the safety scanner SICK 3000. 
It was found that our result lies in the same order of magnitude as stated by SICK. 
In fact, a discrepancy of $4\%$ could be verified. 
However, the data sheet corresponds to a larger scanning range which could explain the difference. 
On the other hand, the statistical significance of our dataset might be one origin for the deviation. 
As a next step, we performed our analysis on data gathered by the use of OpenPose on two datasets. 
We applied OpenPose first on Action Recognition NTU RGB+D, considering the uncertainty behavior of Microsoft Kinect V2. 
In addition, the human pose detection was performed on our data collection. 
Since we made use of three Intel RealSense D435 devices, we considered their uncertainties in our calculations.   
For the analysis, we took the frame rates into account. 
The time-invariant, constant distance between the human joints was defined as conservation measure. 
Conducting statistical analyses, a correlation of 0.32 between the uncertainty and the relative velocity between human and robot could be identified. 
Furthermore, we showed that OpenPose is not suitable for safe HRC: 
According to our results, the uncertainty of OpenPose exceeds the limit required in ISO 13849 by two orders of magnitude. 
In the context of future works, we suggest the incorporation of measurement uncertainties in the robot control framework. 
Uncertainties estimated online could be used to allow for the generation of safe trajectories.  
As the main purpose of our experiments was to demonstrate the applicability of our method, we performed analyses only for short video sequences (max. $30\,s$). 
For a more thorough uncertainty evaluation with a higher statistical significance, we suggest to take more data and use cases into account.

\section{Conclusion and Outlook}
We presented an online uncertainty calculation method that combines a-priori knowledge and system specific data collected at run-time. 
The key idea of our approach lies in the quantification of measurement uncertainties based on system-specific conservation properties. 
Our method allows the online calculation of uncertainties on a user-selected confidence level for customizable attributes. 
Throughout this work, we referred to fundamentals for uncertainty calculation provided in \cite{GUM2009}. 
To do so, we assumed the independence of measurement quantities. 
In future, we aim to consider more complex relationships between the parameters. 
Therefore, we will extend the presented work by Monte Carlo simulations to cover functional behaviors of different uncertainty sources which can be not modeled analytically. 
In the last Section, we explored the uncertainty behavior of OpenPose which is a tool for human position detection. 
By identifying violations of pre-defined conservation properties, we quantified uncertainties of the spatial human detection performance.  
In contrast to sensor devices with manufacturer specifications providing uncertainty information, OpenPose lacks on uncertainty values such that quantifying them is necessary to enable evaluations with respect to safety requirements. 
Furthermore, the incorporation of uncertainties into robot control was not addressed within the scope of this work. 
Assuming the real-time character of robot control, implementing uncertainties arising from the human movement into control algorithms could offer new possibilities in realizing safe and efficient HRC.

\bibliographystyle{IEEEtran}
\bibliography{references}



\end{document}